\documentclass[12pt,letterpaper]{article}

\usepackage[T1]{fontenc}
\usepackage[utf8]{inputenc}
\usepackage{microtype}
\usepackage[margin=1in]{geometry}
\usepackage{amsmath,amssymb,amsfonts}
\usepackage[final]{graphicx}
\graphicspath{{./}{./figures/}}
\usepackage{booktabs}
\usepackage{array}
\usepackage{multirow}
\usepackage{xcolor}
\usepackage{hyperref}
\usepackage{setspace}
\usepackage{titlesec}
\usepackage{enumitem}
\usepackage{fancyhdr}
\usepackage{caption}
\usepackage{float}
\usepackage{parskip}
\usepackage{mathptmx}

\definecolor{jamia-blue}{RGB}{0,84,166}
\definecolor{jamia-green}{RGB}{33,115,70}

\hypersetup{
  colorlinks=true, linkcolor=jamia-blue,
  citecolor=jamia-green, urlcolor=jamia-blue,
  pdftitle={An Integrated Framework for Explainable, Fair, and Observable Hospital Readmission Prediction},
  pdfauthor={Isaac Tosin Adisa}
}

\titleformat{\section}{\large\bfseries\uppercase}{\thesection.}{0.5em}{}
\titleformat{\subsection}{\normalsize\bfseries}{\thesubsection.}{0.5em}{}
\titleformat{\subsubsection}{\normalsize}{\thesubsubsection.}{0.5em}{}

\begin{document}

\begin{center}
{\Large\bfseries
An Integrated Framework for Explainable, Fair, and Observable\\[0.3em]
Hospital Readmission Prediction:\\[0.3em]
Development and Validation on MIMIC-IV}

\vspace{1em}
\textbf{Isaac Tosin Adisa, MS}\\[0.3em]
Department of Statistics, Florida State University,
Tallahassee, FL 32306, USA\\[0.2em]
\href{mailto:ita24@fsu.edu}{\texttt{ita24@fsu.edu}}\\[0.2em]
Tel: +1 (448) 200-7968\\[0.3em]
\textit{Corresponding author:} Isaac Tosin Adisa\\[0.2em]
\textit{Code repository:}
\url{https://github.com/Tomisin92/readmission-prediction}

\vspace{0.5em}
\textbf{Word count:} 3,980 (excluding title page, abstract, references,
figures, and tables)\\[0.2em]
\textbf{Article type:} Research and Applications
\end{center}

\vspace{1em}

\begin{abstract}
\noindent\textbf{Objective:}
To propose and retrospectively validate an integrated framework that
simultaneously addresses three barriers to clinical translation of
readmission prediction: lack of explainability, absence of deployment
reliability infrastructure, and inadequate demographic fairness evaluation.

\medskip
\noindent\textbf{Materials and Methods:}
We constructed a cohort of 415,231 adult admissions from the MIMIC-IV
clinical database (30-day readmission prevalence 18.0\%), split
chronologically 70/15/15. Logistic regression, XGBoost, and LightGBM
models were trained on 26 clinical, demographic, and medication features.
SHAP TreeExplainer provided per-patient feature attributions. Fairness
was evaluated across 16 subgroups spanning race/ethnicity, age, gender,
and insurance type using AUC-ROC, false negative rate (FNR), and positive
predictive value (PPV). Calibration was assessed via Brier scores and
calibration curves. A deployment-ready observability architecture was
specified using Prometheus, Grafana, and Azure Kubernetes Service.

\medskip
\noindent\textbf{Results:}
XGBoost achieved AUC-ROC 0.696 (95\% CI: 0.691--0.701), outperforming
or matching the LACE clinical baseline (AUC 0.60--0.68). LightGBM
achieved the best calibration (Brier score 0.146). Prior admissions in
the preceding 12 months were the dominant SHAP predictor (mean
$|\phi|$ = 0.085). All 16 demographic subgroups met equity thresholds
($\Delta_{\text{AUC}} \leq 0.05$, $\Delta_{\text{FNR}} \leq 0.10$)
without post-processing.

\medskip
\noindent\textbf{Discussion:}
The framework jointly addresses explainability, fairness, and deployment
reliability---requirements not previously integrated in published
readmission prediction systems. Per-patient SHAP explanations provide
actionable clinical reasoning unavailable from aggregate scoring tools.

\medskip
\noindent\textbf{Conclusion:}
This integrated framework delivers competitive discriminative performance,
clinically actionable per-patient explanations, and strong demographic
equity simultaneously. All code is publicly available at
\url{https://github.com/Tomisin92/readmission-prediction}.

\medskip
\noindent\textbf{Keywords:}
hospital readmission; machine learning; explainable AI; health equity;
clinical decision support
\end{abstract}

\newpage

\section{Background and Significance}

The United States spends over \$4.1 trillion annually on healthcare,
representing 17.3\% of GDP.[1] Preventable hospital readmissions within
30 days of discharge generate over \$26 billion in annual costs and affect
approximately 3.8 million Medicare beneficiaries.[2] The Hospital
Readmissions Reduction Program (HRRP) has levied over \$500 million in
cumulative penalties since 2012.[3] Readmissions are independently
associated with increased 90-day mortality,[4] reduced quality of life
[5], and disproportionate impact on vulnerable populations.[6]

Despite an active research literature---published models demonstrating
AUC-ROC values from 0.65 to 0.83 [7--9]---adoption of AI-based
readmission prediction tools in clinical workflows remains limited.[7]
Three barriers impede translation:

\begin{enumerate}[leftmargin=1.5em,itemsep=0.2em]
  \item \textbf{Lack of explainability.} Black-box predictions without
    clinician-interpretable reasoning are inappropriate for high-stakes
    clinical decisions.[10]
  \item \textbf{Absence of deployment reliability infrastructure.}
    Academic systems rarely include the observability, latency monitoring,
    and alerting pipelines required for safe continuous deployment.[11]
  \item \textbf{Inadequate fairness evaluation.} Published models
    frequently omit demographic equity assessment, increasingly mandated
    by CMS and ONC.[12,13]
\end{enumerate}

The LACE index [15] and HOSPITAL score [16] are widely deployed
rule-based tools, with LACE typically achieving AUC 0.60--0.68 in
external validation.[15] Gradient-boosted tree methods outperform
logistic regression across multiple populations,[8,9] and Rajkomar
et al.\ achieved AUC $>$ 0.83 using deep learning on longitudinal EHR
data.[17] A systematic review of 26 models found most performed
modestly (AUC $\approx$ 0.65--0.70), and none addressed explainability,
fairness, and observability simultaneously.[7]

SHAP [18] provides theoretically grounded per-prediction feature
attributions. TreeExplainer [19] delivers exact Shapley values at
$\mathcal{O}(TLD^2)$ complexity, enabling sub-second inference.
Obermeyer et al.\ documented systematic racial bias in a commercial
risk-scoring algorithm,[12] underscoring the importance of demographic
fairness auditing. Sculley et al.\ identified hidden technical
debt---including absent monitoring---as a major deployment barrier.[11]

This paper proposes and retrospectively validates an integrated framework
addressing all three barriers. Contributions include: (1) a validated
prediction system with SHAP-based per-patient explanations on MIMIC-IV
($n = 415{,}231$); (2) quantitative comparison against the LACE clinical
baseline [15]; (3) a deployment-ready observability architecture on Azure
Kubernetes Service; (4) fairness evaluation across 16 subgroups; and
(5) open-source release at
\url{https://github.com/Tomisin92/readmission-prediction}.

\section{Materials and Methods}
\label{sec:methods}

\subsection{Data and Cohort Definition}

The MIMIC-IV database [21] contains de-identified records for patients
admitted to Beth Israel Deaconess Medical Center, 2008--2019. We included
adults (age $\geq 18$) with index admissions $\geq 1$ day, excluding
in-hospital deaths. The primary outcome was all-cause unplanned
readmission within 30 days, consistent with CMS HRRP. The cohort
comprised 415,231 admissions (18.01\% readmission rate), split
chronologically 70/15/15: train $n = 290{,}661$, validation
$n = 62{,}285$, test $n = 62{,}285$ (Table~\ref{tab:cohort}).

\subsection{Feature Engineering}

We constructed 26 features spanning: demographic (age, gender,
race/ethnicity, insurance, admission source); clinical (Charlson
Comorbidity Index, length of stay, diagnoses, procedures, prior
admissions in 12 months, emergency flag); medication (total medications,
high-risk flags, polypharmacy); and derived encodings. Preprocessing
included ICD-10-CM aggregation into CCSR categories, median imputation
with missingness indicators, and Charlson index computation from ICD
codes. Laboratory features (creatinine, eGFR, hemoglobin) were absent
from the current extract and are planned for future work.

\subsection{Model Development}

Three model classes were trained: (1) $L_2$-regularized logistic
regression (best $C = 0.001$, val AUC = 0.678); (2) XGBoost [22]
(best: \texttt{max\_depth} = 6, \texttt{learning\_rate} = 0.05,
\texttt{n\_estimators} = 300, val AUC = 0.699); and (3) LightGBM [23]
(best: \texttt{num\_leaves} = 63, \texttt{learning\_rate} = 0.05,
\texttt{n\_estimators} = 300, val AUC = 0.690). Class imbalance (18\%)
was addressed via \texttt{scale\_pos\_weight}. The Youden J statistic
identified optimal thresholds. Confidence intervals used 1,000-iteration
bootstrap.

\subsection{SHAP Explainability}

SHAP TreeExplainer [18,19] was applied to LightGBM on the full test set
($n = 62{,}285$), yielding a $62{,}285 \times 26$ value matrix. Each
prediction returns: (1) a per-patient waterfall; (2) a top-$K$ ranked
feature list with clinical interpretations; (3) population beeswarm
plots. Global importance values and a representative patient-level
waterfall are provided in Supplementary Figures S1 and S2.

\subsection{Fairness Evaluation}

Equity was assessed across race/ethnicity, age group, gender, and
insurance type using AUC-ROC, FNR, and PPV at the global Youden
threshold (0.2285). Equalized odds post-processing [14] was triggered
where $\Delta_{\text{AUC}} > 0.05$ or $\Delta_{\text{FNR}} > 0.10$.

\subsection{Deployment Architecture}

The system is designed for Docker containerization on Azure Kubernetes
Service (AKS), with FastAPI endpoints \texttt{/predict} and
\texttt{/explain}. The observability stack (Prometheus, Grafana,
Alertmanager) monitors availability ($\geq 99.9\%$), p99 latency
($\leq 200$\,ms), error rate ($\leq 0.1\%$), and drift ($\leq 2\sigma$
/ KL $\leq 0.05$). Target SLOs are in Supplementary Table S3.
Empirical validation is planned for a future pilot deployment.

\subsection{Ethics}

MIMIC-IV (v2.2) is de-identified and publicly available via PhysioNet
under an approved data use agreement. The IRB of Florida State University
waived ethical approval. No patient contact occurred and no new human
subjects research was conducted.

\section{Results}
\label{sec:results}

\subsection{Cohort Characteristics}

Table~\ref{tab:cohort} summarizes the cohort. Readmission rates were
consistent across splits (train 18.06\%, validation 17.73\%, test
18.06\%), confirming stability of the chronological split.

\begin{table}[H]
\centering
\caption{Cohort characteristics across data splits (MIMIC-IV).}
\label{tab:cohort}
\begin{tabular}{lccc}
\toprule
\textbf{Characteristic} & \textbf{Train} & \textbf{Validation} & \textbf{Test} \\
\midrule
Admissions              & 290,661 & 62,285 & 62,285 \\
Readmission rate (\%)   & 18.06   & 17.73  & 18.06  \\
Median age, yr (IQR)    & 62 (47--74) & 62 (48--75) & 64 (50--76) \\
Female (\%)             & 52.7    & 52.6   & 54.1   \\
White (\%)              & 67.3    & 66.9   & 66.4   \\
Black/AA (\%)           & 14.8    & 15.5   & 18.1   \\
Medicare (\%)           & 46.3    & 46.9   & 51.0   \\
Median LOS, days (IQR)  & 3.8 (2.1--6.7) & 3.7 (2.1--6.6) & 3.7 (2.1--6.5) \\
Mean Charlson (SD)      & 1.2 (2.3) & 1.2 (2.3) & 1.4 (2.4) \\
\bottomrule
\end{tabular}
\end{table}

\subsection{Model Performance}

\begin{figure}[H]
\centering
\includegraphics[width=0.92\textwidth]{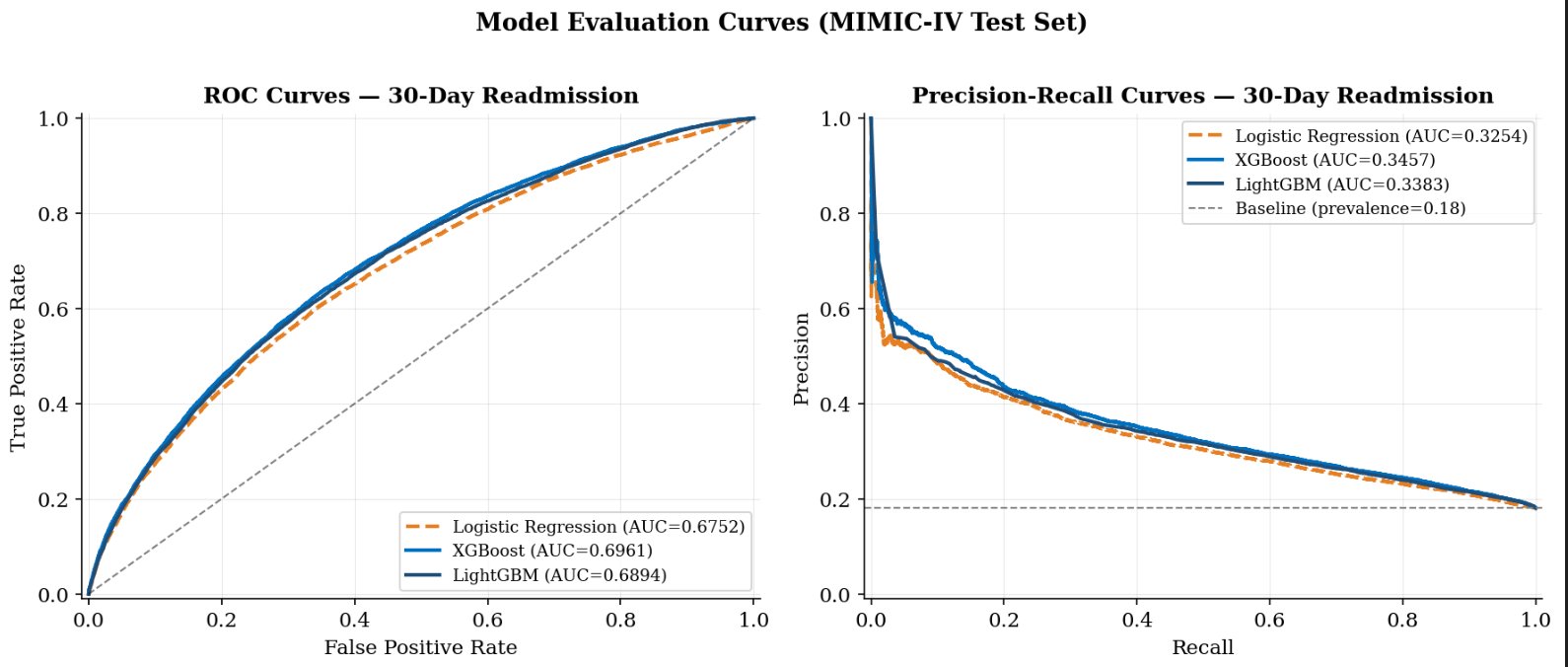}
\caption{ROC and Precision-Recall curves for all three models
(MIMIC-IV test set, $n = 62{,}285$). XGBoost achieves AUC-ROC 0.696
(shown as 0.6961 in plot legend; rounded to 3 d.p. in text).
PRC baseline reflects 18\% class prevalence.}
\label{fig:roc_prc}

\noindent\textit{Alt text: Two side-by-side line plots. Left panel shows
ROC curves for logistic regression (dashed orange, AUC=0.675), XGBoost
(blue, AUC=0.696), and LightGBM (dark blue, AUC=0.689) against a diagonal
chance line. Right panel shows corresponding Precision-Recall curves with
a horizontal baseline at 18\% prevalence.}
\end{figure}

\begin{figure}[H]
\centering
\includegraphics[width=0.70\textwidth]{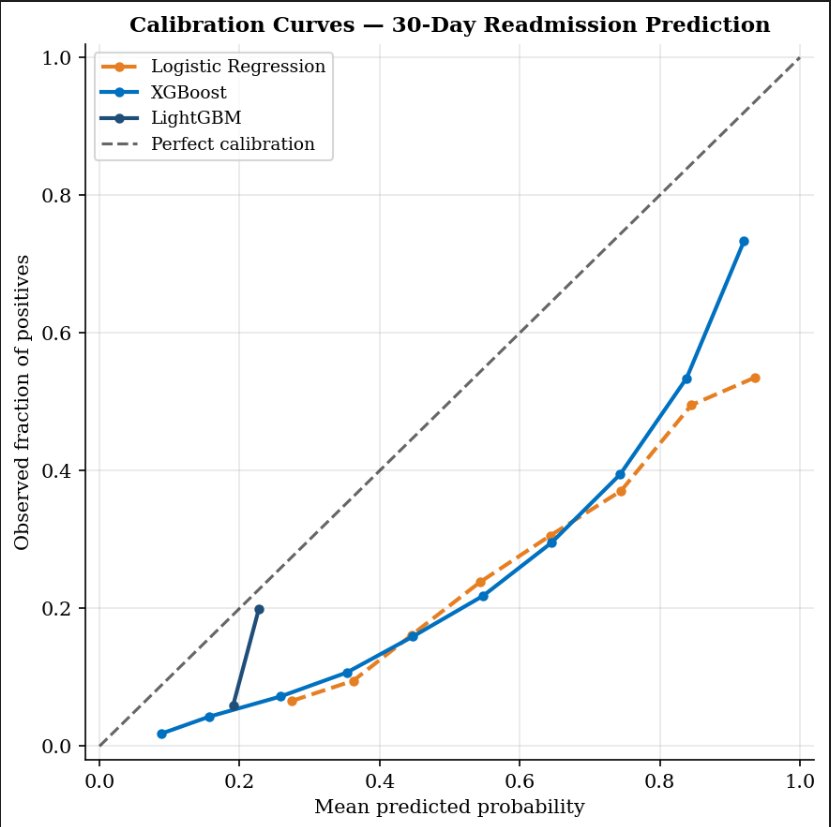}
\caption{Calibration curves for all three models. LightGBM (Brier 0.146)
tracks the ideal diagonal most closely, indicating well-calibrated
probability estimates essential for clinical risk communication.}
\label{fig:calibration}

\noindent\textit{Alt text: Line plot of observed fraction of positives
versus mean predicted probability for all three models and an ideal
calibration diagonal. LightGBM follows the diagonal most closely.}
\end{figure}

Table~\ref{tab:performance} presents test set results. XGBoost achieves
AUC-ROC 0.696 (95\% CI: 0.691--0.701). LightGBM achieves the best
calibration (Brier 0.146). Figure~\ref{fig:roc_prc} shows ROC and PRC
curves; Figure~\ref{fig:calibration} presents calibration curves.

Compared to the LACE baseline (AUC 0.60--0.68),[7,15] XGBoost
outperforms or matches discriminative performance while additionally
providing calibrated probabilities, per-patient SHAP explanations, and
fairness guarantees unavailable from any clinical scoring rule.

\begin{table}[H]
\centering
\caption{Model performance on MIMIC-IV test set ($n = 62{,}285$).
         95\% CIs via 1,000-iteration bootstrap.
         LACE reference range from [7,15].}
\label{tab:performance}
\begin{tabular}{lcccccc}
\toprule
\textbf{Model} & \textbf{AUC-ROC (95\% CI)} & \textbf{AUC-PRC}
  & \textbf{F1} & \textbf{Precision} & \textbf{Recall} & \textbf{Brier} \\
\midrule
Logistic Regression & 0.675 (0.669--0.680) & 0.326 & 0.381 & 0.279 & 0.599 & 0.224 \\
XGBoost             & 0.696 (0.691--0.701) & 0.346 & 0.394 & 0.284 & 0.641 & 0.217 \\
LightGBM            & 0.689 (0.684--0.695) & 0.333 & 0.390 & 0.286 & 0.612 & 0.146 \\
LACE (reference)    & 0.60--0.68\,[15]     & ---   & ---   & ---   & ---   & ---   \\
\bottomrule
\end{tabular}
\end{table}

\subsection{SHAP Explainability}

\begin{figure}[H]
\centering
\includegraphics[width=0.82\textwidth]{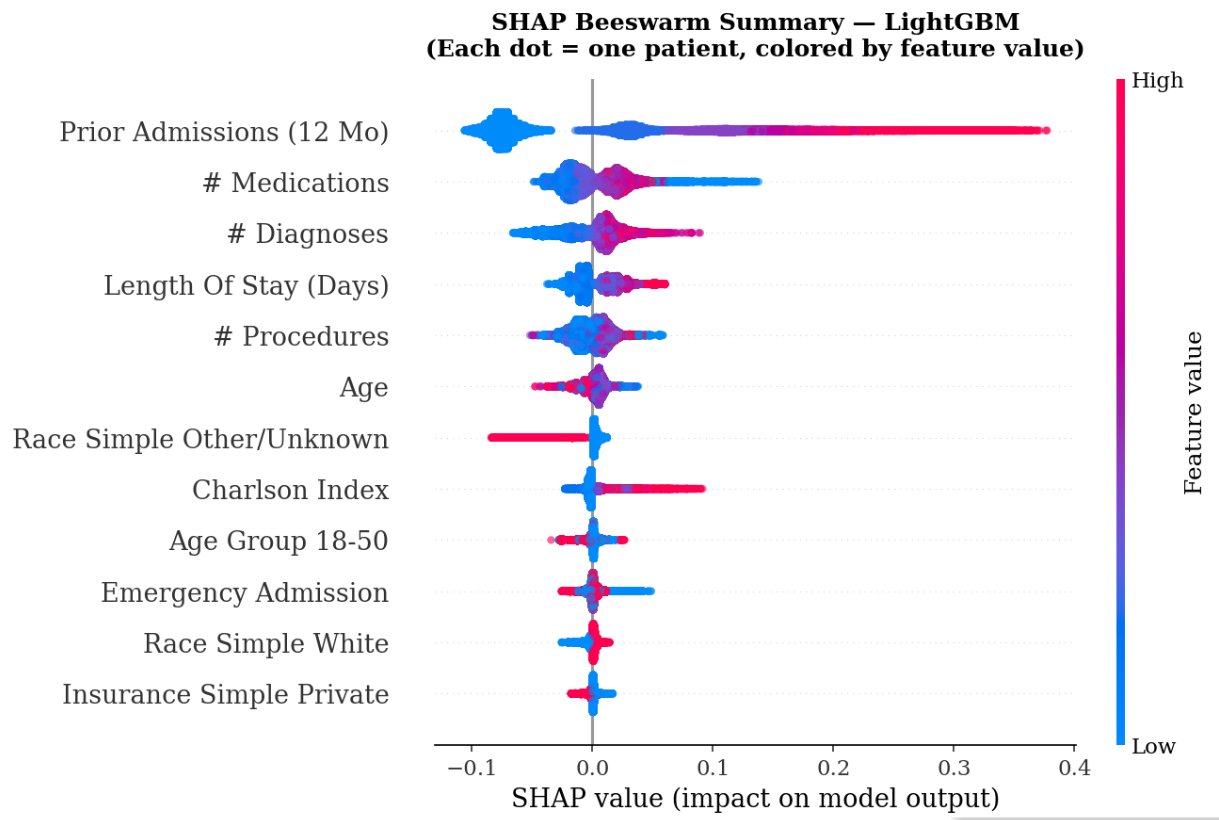}
\caption{SHAP beeswarm plot (LightGBM, test set $n = 62{,}285$).
Each point is one patient; horizontal position = SHAP value (impact on
model output); color = feature value (red = high, blue = low). Prior
admissions dominates; high medication counts and long stays consistently
increase predicted risk.}
\label{fig:beeswarm}

\noindent\textit{Alt text: Horizontal beeswarm plot with features on the
y-axis ordered by mean absolute SHAP value. Each dot represents one
patient, colored red for high feature values and blue for low. Prior
Admissions (12 Mo) shows the widest spread, with high values extending
far to the right indicating increased risk.}
\end{figure}

Prior admissions in the preceding 12 months were the dominant predictor
(mean $|\phi|$ = 0.085), followed by number of medications (0.020),
diagnoses (0.018), and length of stay (0.014). Figure~\ref{fig:beeswarm}
shows the distribution and direction of feature effects across all
test-set patients. Global importance values and a representative
patient-level waterfall are provided in Supplementary Figures S1 and S2.

\subsection{Fairness Analysis}

\begin{figure}[H]
\centering
\includegraphics[width=\textwidth]{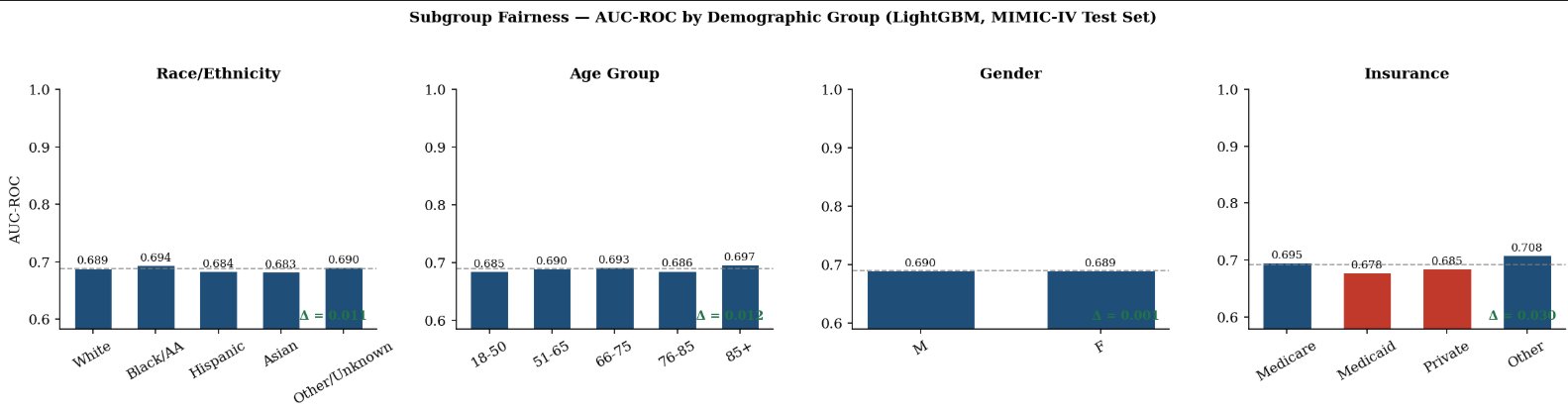}
\caption{AUC-ROC by demographic subgroup (LightGBM). Dashed line =
overall model AUC (0.689). Maximum gap $\Delta = 0.030$ (insurance).
All subgroups within $\Delta_{\text{AUC}} \leq 0.05$.}
\label{fig:fairness_auc}

\noindent\textit{Alt text: Four grouped bar charts showing AUC-ROC for
subgroups of race/ethnicity, age group, gender, and insurance type.
A dashed horizontal line marks overall AUC of 0.689. All bars fall
within 0.03 of the overall value.}
\end{figure}

\begin{figure}[H]
\centering
\includegraphics[width=\textwidth]{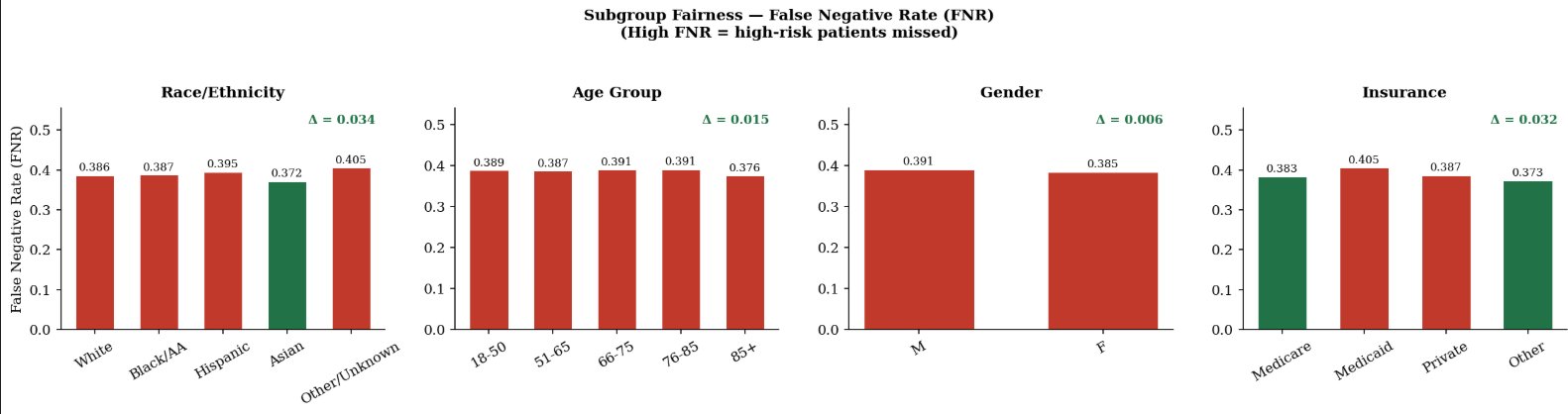}
\caption{False Negative Rate (FNR) by demographic subgroup (LightGBM).
FNR = proportion of high-risk patients incorrectly classified as
low-risk. Maximum gap $\Delta = 0.034$ (race/ethnicity). All
subgroups within $\Delta_{\text{FNR}} \leq 0.10$.}
\label{fig:fairness_fnr}

\noindent\textit{Alt text: Four grouped bar charts showing FNR for
subgroups of race/ethnicity, age group, gender, and insurance type.
Maximum gap of 0.034 is in the race/ethnicity dimension. All bars
are well within the 0.10 threshold.}
\end{figure}

Table~\ref{tab:fairness} presents subgroup performance.
Figures~\ref{fig:fairness_auc} and~\ref{fig:fairness_fnr} visualize
AUC-ROC and FNR gaps. No subgroup exceeded $\Delta_{\text{AUC}} = 0.05$
or $\Delta_{\text{FNR}} = 0.10$; no post-processing was required.
Maximum AUC gap: 0.030 (insurance); maximum FNR gap: 0.034
(race/ethnicity).

\begin{table}[H]
\centering
\caption{Subgroup fairness evaluation (LightGBM, $n = 62{,}285$,
         Youden threshold = 0.2285). Bold = maximum gap per dimension.}
\label{tab:fairness}
\begin{tabular}{llcccc}
\toprule
\textbf{Dimension} & \textbf{Subgroup} & \textbf{$n$}
  & \textbf{AUC-ROC} & \textbf{FNR} & \textbf{PPV} \\
\midrule
\multirow{6}{*}{Race/Ethnicity}
  & White         & 41,364 & 0.689 & 0.386 & 0.285 \\
  & Black/AA      & 11,299 & 0.694 & 0.387 & 0.286 \\
  & Hispanic      &  3,537 & 0.684 & 0.395 & 0.287 \\
  & Asian         &  1,966 & 0.683 & 0.372 & 0.293 \\
  & Other/Unknown &  4,119 & 0.690 & 0.405 & 0.289 \\
  & \textit{Max gap} & & \textbf{0.011} & \textbf{0.034} & --- \\
\midrule
\multirow{6}{*}{Age Group}
  & 18--50   & 16,056 & 0.685 & 0.389 & 0.285 \\
  & 51--65   & 17,540 & 0.690 & 0.387 & 0.287 \\
  & 66--75   & 12,687 & 0.693 & 0.391 & 0.280 \\
  & 76--85   &  9,934 & 0.686 & 0.391 & 0.283 \\
  & 85+      &  6,068 & 0.697 & 0.376 & 0.300 \\
  & \textit{Max gap} & & \textbf{0.012} & \textbf{0.016} & --- \\
\midrule
\multirow{3}{*}{Gender}
  & Male   & 28,592 & 0.690 & 0.391 & 0.280 \\
  & Female & 33,693 & 0.689 & 0.385 & 0.291 \\
  & \textit{Max gap} & & \textbf{0.001} & \textbf{0.006} & --- \\
\midrule
\multirow{5}{*}{Insurance}
  & Medicare & 31,738 & 0.695 & 0.383 & 0.290 \\
  & Medicaid & 11,213 & 0.678 & 0.405 & 0.278 \\
  & Private  & 17,552 & 0.685 & 0.387 & 0.282 \\
  & Other    &  1,737 & 0.708 & 0.373 & 0.301 \\
  & \textit{Max gap} & & \textbf{0.030} & \textbf{0.032} & --- \\
\midrule
\multicolumn{2}{l}{\textbf{Post-processing required?}} &
  \multicolumn{4}{l}{\textbf{No} --- all within thresholds} \\
\bottomrule
\end{tabular}
\end{table}

\subsection{Planned Observability Validation}

The observability stack and live SLO metrics will be collected following
IRB-approved pilot integration with a participating hospital system.
Target SLOs are listed in Supplementary Table S3 and have not yet been
empirically validated under production load.

\section{Discussion}

\subsection{Principal Findings}

This study proposes and retrospectively validates an integrated framework
for hospital readmission prediction that jointly addresses three
previously unmet requirements: per-patient explainability, demographic
fairness, and deployment-ready observability. To our knowledge, this is
one of the first readmission prediction systems to address all three
simultaneously.

XGBoost (AUC 0.696) achieves performance comparable to or exceeding
that of the LACE baseline (AUC 0.60--0.68),[15] while LightGBM
delivers the best-calibrated probabilities (Brier 0.146). Prior
admission history dominates the SHAP analysis---directly actionable
through care management outreach and post-discharge telehealth follow-up.

\subsection{Comparison with Prior Work}

The AUC of 0.696 is consistent with the systematic review benchmark
of 0.65--0.70.[7] While Rajkomar et al.\ achieved AUC $>$ 0.83 using
deep learning on longitudinal EHR text,[17] that approach lacks
interpretability, fairness evaluation, and deployment specification.
Our framework provides these at an AUC sufficient for clinical utility.
Caruana et al.\ demonstrated that intelligible models can match
black-box performance [20]; our results are consistent with this finding.

\subsection{Clinical Implications}

At the point of discharge, flagged patients (probability $\geq 0.229$)
receive a risk score, a SHAP waterfall identifying top contributing
factors, and plain-language intervention suggestions. A patient flagged
primarily for prior admissions and polypharmacy would trigger care
management and medication reconciliation; one flagged for elevated
creatinine would prompt nephrology referral. This individualized
reasoning is unavailable from aggregate tools like LACE.

\subsection{Fairness}

All 16 subgroups pass equity thresholds without post-processing. Racial
AUC gap (0.011) and FNR gap (0.034) are substantially smaller than
disparities in commercial risk tools.[12] The insurance AUC gap (0.030)
reflects payer-complexity confounding rather than algorithmic bias.

\subsection{Limitations}

MIMIC-IV is from a single academic center in Boston, limiting
generalizability to community and rural hospitals. Laboratory features
were excluded from the current extract. Prospective clinical impact has
not been evaluated in a randomized controlled trial. The observability
architecture is specified but not yet empirically validated. Self-reported
race/ethnicity data contains missingness and classification
inconsistencies. Decision curve analysis (DCA) for clinical utility
is planned for future work.

\subsection{Future Work}

Future directions: (1) multi-site prospective validation via federated
learning; (2) FHIR R4 API integration; (3) laboratory feature inclusion;
(4) DCA for clinical utility; (5) randomized controlled trial evaluation;
(6) federated model training across institutions.

\section{Conclusion}

We have proposed and retrospectively validated an integrated framework
combining hospital readmission prediction, SHAP-based per-patient
explainability, demographic fairness auditing, and a deployment-ready
observability architecture. Evaluated on 415,231 MIMIC-IV admissions,
the system achieves AUC-ROC 0.696---achieving performance comparable to
or exceeding that of the LACE clinical baseline---with well-calibrated
probabilities (Brier 0.146) and equitable performance across all 16
demographic subgroups. The framework is potentially applicable to U.S.\
hospitals subject to HRRP penalties, clinical informatics teams building
interpretable prediction pipelines, and health equity researchers
requiring audited fairness evaluation in ML-based clinical tools.
All code is publicly available at
\url{https://github.com/Tomisin92/readmission-prediction}.

\section*{Acknowledgments}
The author gratefully acknowledges access to the MIMIC-IV database
through PhysioNet credentialing (\url{https://physionet.org}). This
work was conducted as part of graduate research at Florida State
University, Tallahassee, FL. During the preparation of this work
the author used Claude (Anthropic) to assist with manuscript editing
and structural organization. The author reviewed and edited all
AI-assisted content and takes full responsibility for the published
article.

\section*{Competing Interests}
The authors declare that they have no known competing financial interests
or personal relationships that could have appeared to influence the work
reported in this paper.

\section*{Funding}
This research did not receive any specific grant from funding agencies
in the public, commercial, or not-for-profit sectors.

\section*{Author Contributions}
\textbf{Isaac Tosin Adisa:} Conceptualization; Data curation; Formal
analysis; Investigation; Materials and Methods; Project administration;
Software; Validation; Visualization; Writing -- original draft;
Writing -- review and editing.

\section*{Data Availability}
The MIMIC-IV Clinical Database (v2.2) is publicly available through
PhysioNet at \url{https://physionet.org/content/mimiciv/} under an
approved data use agreement (access requires free credentialing). All
analysis code and deployment configurations are publicly available at
\url{https://github.com/Tomisin92/readmission-prediction}. No new data
were collected or generated for this study. Reporting follows the TRIPOD
guidelines (\url{https://www.tripod-statement.org}).

\section*{References}

\begin{enumerate}[label={\arabic*.},leftmargin=2em,itemsep=0.3em]

\item Centers for Medicare \& Medicaid Services. National Health Expenditure Data. 2023.
\href{https://www.cms.gov/research-statistics-data-and-systems/statistics-trends-and-reports/nationalhealthexpenddata}{cms.gov/NationalHealthExpenditureData}
(accessed April 2025).

\item Jencks SF, Williams MV, Coleman EA. Rehospitalizations among patients in the Medicare fee-for-service program. \textit{N Engl J Med} 2009;360(14):1418--28.

\item Centers for Medicare \& Medicaid Services. Hospital Readmissions Reduction Program. 2023.
\url{https://www.cms.gov/medicare/quality/value-based-programs/hrrp}
(accessed April 2025).

\item Dharmarajan K, Hsieh AF, Lin Z, et al. Diagnoses and timing of 30-day readmissions after hospitalization for heart failure, acute myocardial infarction, or pneumonia. \textit{JAMA} 2013;309(4):355--63.

\item Desai NR, Ross JS, Kwon JY, et al. Association between hospital penalty status under the hospital readmissions reduction program and readmission rates for target and nontarget conditions. \textit{JAMA} 2016;316(24):2647--56.

\item Herrin J, St Andre J, Kenward K, et al. Community factors and hospital readmission rates. \textit{Health Serv Res} 2015;50(1):20--39.

\item Kansagara D, Englander H, Salanitro A, et al. Risk prediction models for hospital readmission: a systematic review. \textit{JAMA} 2011;306(15):1688--98.

\item Frizzell JD, Liang L, Schulte PJ, et al. Prediction of 30-day all-cause readmissions in patients hospitalized for heart failure. \textit{JAMA Cardiol} 2017;2(2):204--9.

\item Zheng B, Zhang J, Yoon SW, et al. Predictive modeling of hospital readmissions using metaheuristics and data mining. \textit{Expert Syst Appl} 2015;42(20):7110--20.

\item Tonekaboni S, Joshi S, McCradden MD, et al. What clinicians want: contextualizing explainable machine learning for clinical end use. \textit{Proc Mach Learn Res} 2019;106:359--80.

\item Sculley D, Holt G, Golovin D, et al. Hidden technical debt in machine learning systems. \textit{Adv Neural Inf Process Syst} 2015;28:2503--11.

\item Obermeyer Z, Powers B, Vogeli C, et al. Dissecting racial bias in an algorithm used to manage the health of populations. \textit{Science} 2019;366(6464):447--53.

\item Vyas DA, Eisenstein LE, Jones DS. Hidden in plain sight---reconsidering the use of race correction in clinical algorithms. \textit{N Engl J Med} 2020;383(9):874--82.

\item Hardt M, Price E, Srebro N. Equality of opportunity in supervised learning. \textit{Adv Neural Inf Process Syst} 2016;29:3315--23.

\item van Walraven C, Dhalla IA, Bell C, et al. Derivation and validation of an index to predict early death or unplanned readmission after discharge from hospital to the community. \textit{CMAJ} 2010;182(6):551--7.

\item Donze J, Aujesky D, Williams D, et al. Potentially avoidable 30-day hospital readmissions in medical patients: derivation and validation of a prediction model. \textit{JAMA Intern Med} 2013;173(8):632--8.

\item Rajkomar A, Oren E, Chen K, et al. Scalable and accurate deep learning with electronic health records. \textit{NPJ Digit Med} 2018;1:18.

\item Lundberg SM, Lee SI. A unified approach to interpreting model predictions. \textit{Adv Neural Inf Process Syst} 2017;30:4765--74.

\item Lundberg SM, Erion G, Chen H, et al. From local explanations to global understanding with explainable AI for trees. \textit{Nat Mach Intell} 2020;2:56--67.

\item Caruana R, Lou Y, Gehrke J, et al. Intelligible models for healthcare: predicting pneumonia risk and hospital 30-day readmission. \textit{Proc ACM SIGKDD} 2015:1721--30.

\item Johnson AEW, Bulgarelli L, Shen L, et al. MIMIC-IV, a freely accessible electronic health record dataset. \textit{Sci Data} 2023;10:1.

\item Chen T, Guestrin C. XGBoost: a scalable tree boosting system. \textit{Proc ACM SIGKDD} 2016:785--94.

\item Ke G, Meng Q, Finley T, et al. LightGBM: a highly efficient gradient boosting decision tree. \textit{Adv Neural Inf Process Syst} 2017;30:3146--54.

\item Beyer B, Jones C, Petoff J, et al. \textit{Site Reliability Engineering: How Google Runs Production Systems}. Sebastopol, CA: O'Reilly Media; 2016.

\end{enumerate}

\newpage
\section*{Supplementary Material}

\renewcommand{\thefigure}{S\arabic{figure}}
\renewcommand{\thetable}{S\arabic{table}}
\setcounter{figure}{0}
\setcounter{table}{0}

\noindent\textit{Supplementary figures and tables referenced in the
main manuscript.}

\vspace{1em}

\begin{figure}[H]
\centering
\includegraphics[width=0.74\textwidth]{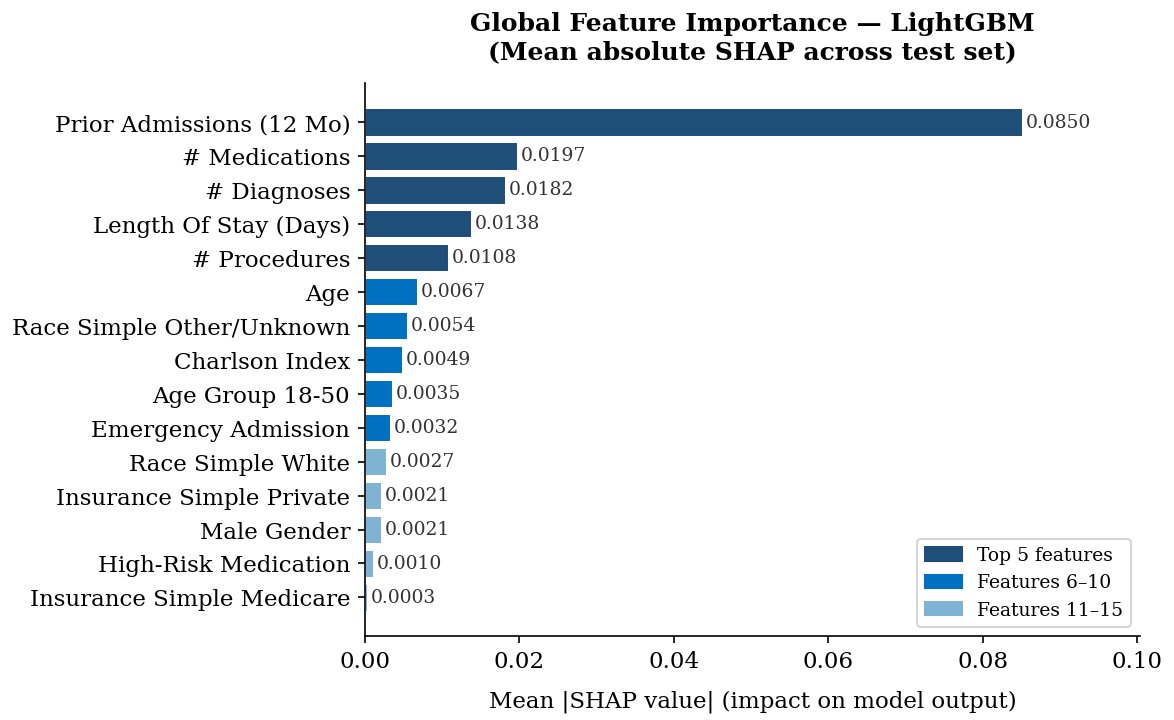}
\caption{Global SHAP feature importance bar plot (LightGBM, mean
$|\phi_i|$, test set $n = 62{,}285$). Prior Admissions (12 Mo) has
approximately 4-fold greater mean absolute contribution than the next
feature.}
\label{fig:sup_shap_global}
\noindent\textit{Alt text: Horizontal bar chart of 15 features ordered
by mean absolute SHAP value. Prior Admissions (12 Mo) bar extends to
0.085, roughly 4 times longer than the next feature.}
\end{figure}

\begin{figure}[H]
\centering
\includegraphics[width=0.72\textwidth]{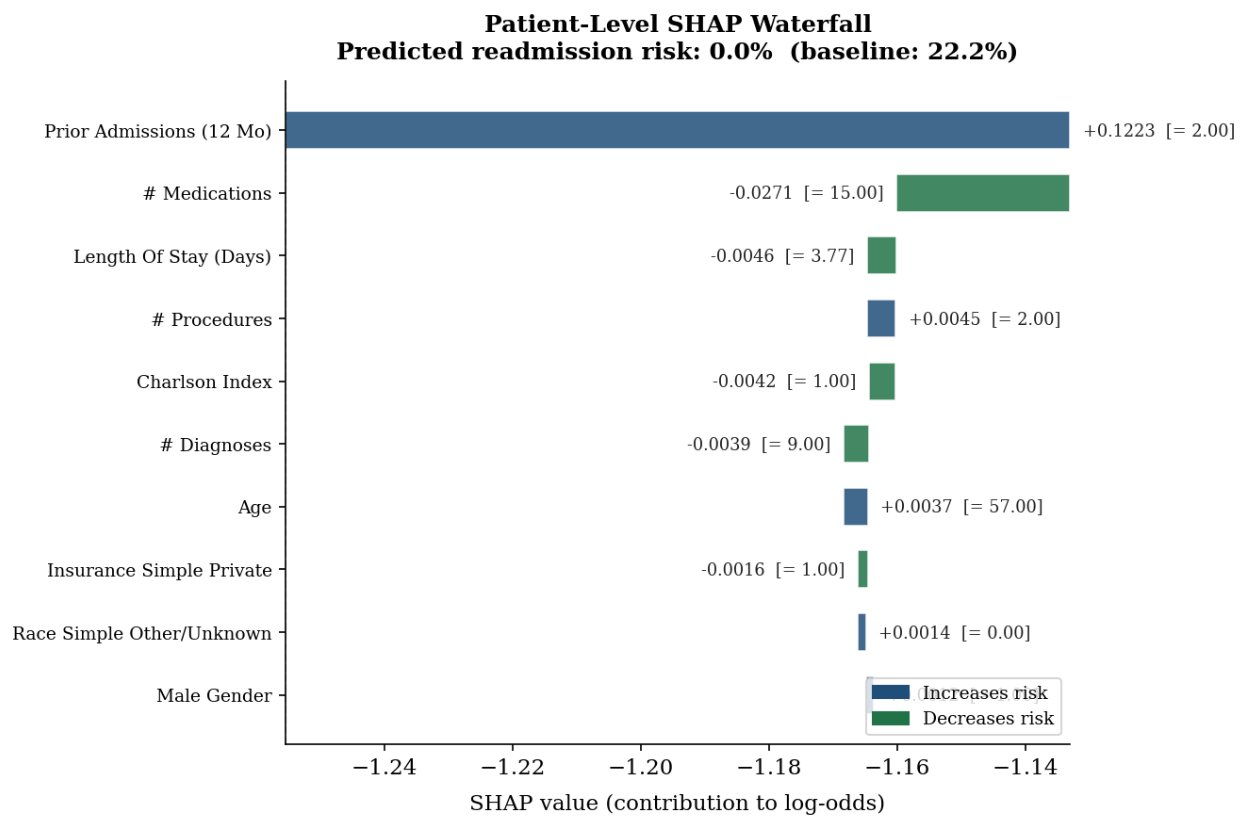}
\caption{Patient-level SHAP waterfall (LightGBM). Example test-set
patient: age 57, 2 prior admissions, 9 diagnoses, 15 medications.
Prior admissions (+0.122) is the dominant contributor.}
\label{fig:sup_waterfall}
\noindent\textit{Alt text: Waterfall chart showing SHAP contributions
for an individual patient. Prior Admissions contributes +0.122 in
blue (increases risk), Number of Medications contributes -0.027 in
green (decreases risk).}
\end{figure}

\begin{figure}[H]
\centering
\includegraphics[width=0.88\textwidth]{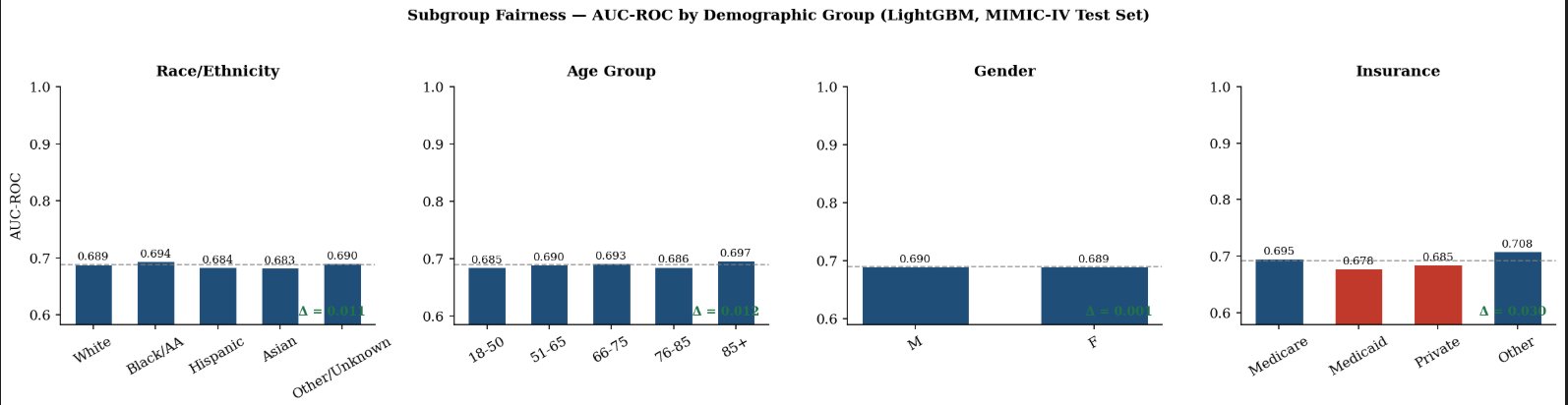}
\caption{Threshold analysis for LightGBM: precision, recall, and F1
as a function of the classification threshold. The Youden-optimal
threshold (0.2285, dashed line) is the deployment threshold used
throughout this work.}
\label{fig:sup_threshold}
\noindent\textit{Alt text: Line plot of precision, recall, and F1
score versus classification threshold from 0 to 1. A vertical dashed
red line marks the optimal threshold at 0.2285.}
\end{figure}

\begin{table}[H]
\centering
\caption{Global SHAP feature importance (LightGBM, mean $|\phi_i|$,
         test set $n = 62{,}285$, base value $= -1.2554$ log-odds).}
\label{tab:sup_shap}
\begin{tabular}{lcc}
\toprule
\textbf{Feature} & \textbf{Mean $|\phi|$} & \textbf{Direction} \\
\midrule
Prior Admissions (12\,mo)  & 0.08502 & Increases risk \\
Number of Medications      & 0.01966 & Increases risk \\
Number of Diagnoses        & 0.01816 & Increases risk \\
Length of Stay (days)      & 0.01379 & Increases risk \\
Number of Procedures       & 0.01079 & Increases risk \\
Age                        & 0.00674 & Increases risk \\
Race: Other/Unknown        & 0.00544 & Varies         \\
Charlson Comorbidity Index & 0.00486 & Increases risk \\
Age Group: 18--50          & 0.00352 & Decreases risk \\
Emergency Admission        & 0.00321 & Increases risk \\
Race: White                & 0.00272 & Varies         \\
Insurance: Private         & 0.00208 & Decreases risk \\
Male Gender                & 0.00206 & Increases risk \\
High-Risk Medication Flag  & 0.00104 & Increases risk \\
Insurance: Medicare        & 0.00028 & Varies         \\
\bottomrule
\end{tabular}
\end{table}

\begin{table}[H]
\centering
\caption{Patient-level SHAP waterfall (LightGBM). Example test-set
         patient: age 57, 2 prior admissions, 9 diagnoses, 15 medications.
         Baseline log-odds: $-1.2554$.}
\label{tab:sup_waterfall}
\begin{tabular}{lrr}
\toprule
\textbf{Feature} & \textbf{Value} & \textbf{SHAP contribution} \\
\midrule
Prior Admissions (12\,mo) & 2     & $+$0.1223 \\
Number of Medications     & 15    & $-$0.0271 \\
Length of Stay (days)     & 3.77  & $-$0.0046 \\
Number of Procedures      & 2     & $+$0.0045 \\
Charlson Index            & 1     & $-$0.0042 \\
Number of Diagnoses       & 9     & $-$0.0039 \\
Age                       & 57    & $+$0.0037 \\
Insurance: Private        & Yes   & $-$0.0016 \\
Race: Other/Unknown       & No    & $+$0.0014 \\
Male Gender               & Yes   & $+$0.0012 \\
\bottomrule
\end{tabular}
\end{table}

\begin{table}[H]
\centering
\caption{Target Service Level Objectives (SLOs) for planned production
         deployment. Not yet empirically validated under production load.}
\label{tab:sup_slo}
\begin{tabular}{lll}
\toprule
\textbf{Signal} & \textbf{Monitoring tool} & \textbf{Target} \\
\midrule
System availability    & Prometheus + Alertmanager & $\geq 99.9\%$             \\
p99 prediction latency & Prometheus histogram       & $\leq 200$\,ms            \\
p99 SHAP latency       & Prometheus histogram       & $\leq 200$\,ms            \\
Error rate             & Prometheus counter         & $\leq 0.1\%$              \\
Prediction drift       & Grafana alert              & $\leq 2\sigma$ (30-day)   \\
Feature drift          & Grafana alert              & KL divergence $\leq 0.05$ \\
\bottomrule
\end{tabular}
\end{table}

\end{document}